\documentclass[final]{cvpr}

\usepackage{times}
\usepackage{epsfig}
\usepackage{graphicx}
\usepackage{amsmath}
\usepackage{amssymb}
\usepackage{float}

\usepackage[pagebackref=true,breaklinks=true,colorlinks,bookmarks=false]{hyperref}

\def\cvprPaperID{AICC} 
\def\confYear{CVPR 2021}

\title{Texture Generation with Neural Cellular Automata}

\author{Alexander Mordvintsev\footnotemark \hspace{10mm} Eyvind Niklasson\footnotemark[\value{footnote}] \hspace{10mm} Ettore Randazzo\\
Google Research\\
{\tt\small \{moralex, eyvind, etr\}@google.com}
}

\begin{document}
\twocolumn[{%
\renewcommand\twocolumn[1][]{#1}%
\maketitle
\begin{center}
\includegraphics[width=0.95\linewidth]{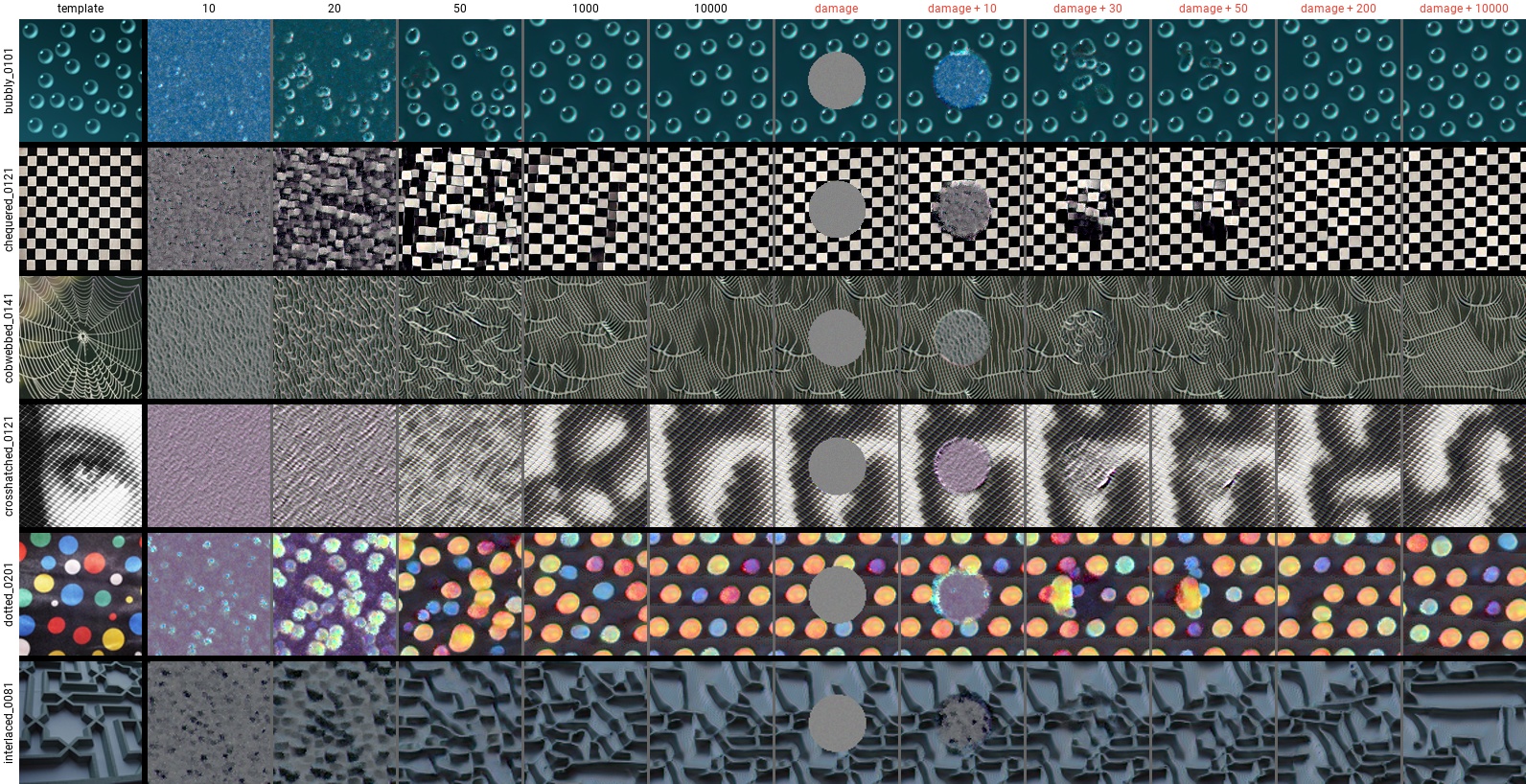}
\end{center}
\label{fig:hero}
}]

\maketitle

\renewcommand*{\thefootnote}{\fnsymbol{footnote}}
\footnotetext{* Contributed equally.}
\renewcommand*{\thefootnote}{\arabic{footnote}}

\begin{abstract}
Neural Cellular Automata (NCA\footnote{We use NCA to mean both Neural Cellular \textbf{Automata} and Neural Cellular \textbf{Automaton} in this work.}) have shown a remarkable ability to learn the required rules to ”grow” images \cite{mordvintsev2020growing}, classify morphologies \cite{randazzo2020self-classifying} and  segment images \cite{sandler2020image}, as well as to do general computation such as path-finding \cite{endo_yasuoka}. We believe the inductive prior they introduce lends itself to the generation of textures. Textures in the natural world are often generated by variants of locally interacting reaction-diffusion systems. Human-made textures are likewise often generated in a local manner (textile weaving, for instance) or using rules with local dependencies (regular grids or geometric patterns). We demonstrate learning a texture generator from a single template image, with the generation method being embarrassingly parallel, exhibiting quick convergence and high fidelity of output, and requiring only some minimal assumptions around the underlying state manifold. Furthermore, we investigate properties of the learned models that are both useful and interesting, such as non-stationary dynamics and an inherent robustness to damage. Finally, we make qualitative claims that the behaviour exhibited by the NCA model is a learned, distributed, local algorithm to generate a texture, setting our method apart from existing work on texture generation. We discuss the advantages of such a paradigm.
\end{abstract}
\section{Introduction}

Texture synthesis is an actively studied problem of computer graphics and image processing. Most of the work in this area is focused on creating new images of a texture specified by the provided image pattern \cite{Efros1999TextureSB,Lefebvre2006AppearancespaceTS}. These images should give the impression, to a human observer, that they are generated by the same stochastic process that generated the provided sample. An alternative formulation of the texture synthesis problem is searching for a stochastic process that allows efficient sampling from the texture image distribution defined by the input image. With the advent of deep neural networks, feed-forward convolutional generators have been proposed that transform latent vectors of random i.i.d. values into texture image samples \cite{Ulyanov2016TextureNF}.

Many texture patterns observed in nature result from local interactions between tiny particles, cells or molecules, which lead to the formation of larger structures. This distributed process of pattern formation is often referred to as self-organisation. Typical computational models of such systems are systems of PDEs \cite{Turing1990TheCB,Chan2020LeniaAE}, cellular automata, multi-agent or particle systems.

In this work, we use the recently proposed Neural Cellular Automata (NCA) \cite{mordvintsev2020growing,randazzo2020self-classifying,sandler2020image} as a biologically plausible model of distributed texture pattern formation. The image generation process is modelled as an asynchronous, recurrent computation, performed by a population of locally-communicating cells arranged in a regular 2D grid. All cells share the same differentiable update rule. We use backpropagation through time and a grid-wide differentiable objective function to train the update rule, which is able to synthesise a pattern similar to a provided example.

The proposed approach achieves a very organic-looking dynamic of progressive texture synthesis through local communication and allows great flexibility in post-training adaptation. The decentralized, homogeneous nature of the computation performed by the learned synthesis algorithm potentially allows the embedding of their implementations in future media, such as smart fabric displays or electronic decorative tiles.

\section{Neural CA image generator}
We base our image generator on the Neural Cellular Automata model \cite{mordvintsev2020growing}. Here we summarize the key elements of the model and place them in the context of PDEs, cellular automata, and neural networks to highlight different features of the model. 

\subsection{Pattern-generating PDE systems}
Systems of partial differential equations (PDEs) have been used to model natural pattern formation processes for a long time. Well known examples include the seminal work by Turing \cite{Turing1990TheCB}, or the Grey-Scott reaction diffusion patterns \cite{Munafo2014StableLM}. It seems quite natural to use PDEs for texture synthesis. Specifically, given a texture image sample, we are looking for a function $f$ that defines the evolution of a vector function $\mathbf{s}(\mathbf{x}, t)$, defined on a two-dimensional manifold $\mathbf{x}$:
$$\frac{\partial \mathbf{s} }{\partial  t } = f(\textbf{s}, \nabla_\mathbf{x} \textbf{s}, \nabla_\mathbf{x}^{2}\textbf{s})$$
where $\mathbf{s}$ represents a $k$ dimensional vector, whose first three components correspond to the visible RGB color channels: $\mathbf{s} = (s^0=R, s^1=G, s^2=B, s^3, ... , s^{k-1})$. The RGB channels should form the texture, similar to the provided example. $\nabla_\mathbf{x} \mathbf{s}$ denotes a matrix of per-component gradients over $\mathbf{x}$, and $\nabla_\mathbf{x}^{2}\textbf{s}$ is a vector of laplacians \footnote{We added the Lapacian kernel to make the system general enough to reproduce the Gray-Scott reaction-diffusion system.}. The evolution of the system starts with some initial state $\mathbf{s}_0$ and is guided by a space-time uniform rule of $f$. We don't imply the existence of a static final state of the pattern evolution, but just want the system to produce an input-like texture as early as possible, and perpetually maintain this similarity .

\subsection{From PDEs to Cellular Automata}
\label{section:ca_model}
In order to evaluate the behaviour of a PDE system on a digital computer, one must discretize the spatio-temporal domain, provide the discrete versions of gradient and Laplacian operators, and specify the integration algorithm. During training we use a uniform Cartesian raster 2D grid with torus topology (i.e. wrap-around boundary conditions). Note that the system now closely mirrors that of a Cellular Automata - there is a uniform raster grid, with each point undergoing time evolution dependant only on the neighbouring cells. The evolution of the CA state $\mathbf{s}_t(x, y)$, where $x$ and $y$ are integer cell coordinates, is now given by
\begin{align*}
\mathbf{p}_{t} &=  concat(\mathbf{s}_{t}, K_{x} \ast \mathbf{s}_t, K_{y} \ast \mathbf{s}_t, K_{lap} \ast \mathbf{s}_t) \\
\mathbf{s}_{t+1} &= \mathbf{s}_{t} + f(\mathbf{p}_{t}) \delta_{x, y, t}
\end{align*}
Discrete approximations of gradient and Laplacian operators are provided by linear convolutions with a set of 3x3 kernels $K_x$, $K_y$ and $K_{lap}$. We use Sobel filters \cite{Sobel1990AnI3} and a 9-point variant of the discrete Laplacian:

\begin{center}
\scalebox{.75}{\begin{math}\begin{array}{ c c c }
\begin{bmatrix}
-1 & 0 & 1\\-2 & 0 & 2 \\-1 & 0 & 1 \\
\end{bmatrix}
&
\begin{bmatrix}
-1 & -2 & -1\\ 0 & 0 & 0 \\1 & 2 & 1 \\
\end{bmatrix}
&
\begin{bmatrix}
1 & 2 & 1\\2 & -12 & 2 \\1 & 2 & 1 \\
\end{bmatrix}
\\
K_x & K_y & K_{lap}
\end{array}
\end{math}}
\end{center}

We call $\mathbf{p}$ a perception vector, as it gathers information about the neighborhood of each cell through convolution kernels. The function $f$ is the per-cell learned update rule that we obtain using the optimisation process, described later. The separation between perception and update rules allows us to transfer learned rules to different grid structures and topologies, as long as the gradient and Laplacian operators are provided (see section \ref{section:post_training}).

\paragraph{Stochastic updates} The cell update rate is denoted by $\delta_{x, y, t}$. In the case of the uniform update rate ($\delta_{x, y, t}=c$), the above rule can be interpreted as a step of the explicit Euler integration method. If all cells are updated synchronously, initial conditions $\mathbf{s}_0$  have to vary from cell-to-cell in order to break the symmetry. This can be achieved by initializing the grid with random noise. The physical implementation of the synchronous model would require existence of a global clock, shared by all cells. In the spirit of self-organisation, we tried to decouple the cell updates. Following the \cite{mordvintsev2020growing}, we emulate\footnote{This is a pretty rough model of the real world asynchronous computation, yet it seems to generalise well into the unforeseen scenarios, like two adjacent grid regions exhibiting very different update rates (fig. \ref{fig:parallel}).} the asynchronous cell updates by independently sampling $\delta_{x, y, t}$ from $\{0, 1\}$ for each cell at each step, with $\Pr(\delta_{x, y, t}=1)=0.5$. Asynchronous updates allow to CA to break the symmetry even for the uniform initial state $s_0$.

\subsection{From CA to Neural Networks}
\begin{figure}[t]
\begin{center}
   \includegraphics[width=0.95\linewidth]{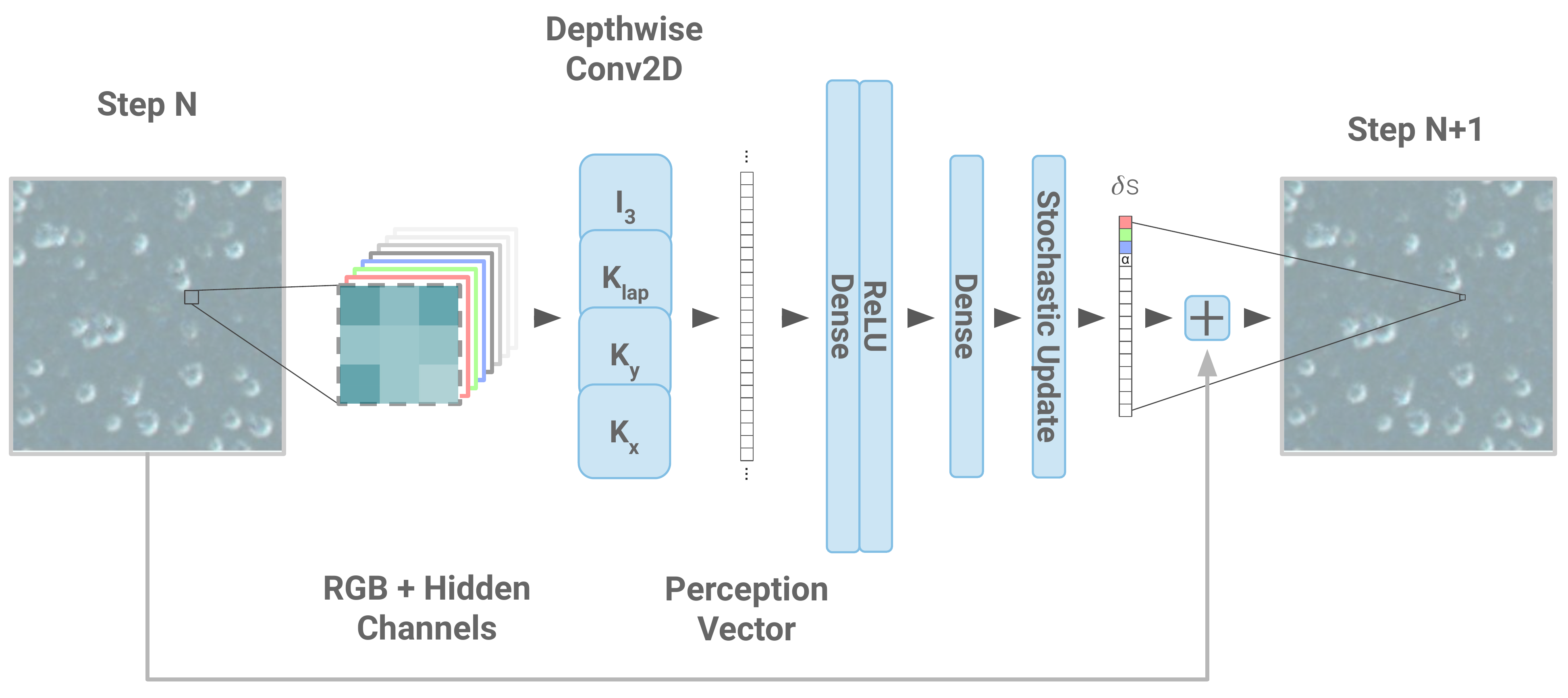}
\end{center}
\caption{Texture NCA model architecture.}
\label{fig:simplified}
\end{figure}

The last component that we have to define is the update function. We use $f(\mathbf{p}) = relu(\mathbf{p} W_0 + b_0) W_1 + b_1$, where $\mathbf{p}$ is a perception vector, and $W_{0,1}$, $b_{0,1}$ are the learned parameters. If we look at the resulting system from the differentiable programming perspective, we can see that the whole CA image generator can be represented by a recurrent convolutional neural network (Fig.\ref{fig:simplified}), that can be built from standard components, available in modern deep learning frameworks. Using the established neural net terminology, we can call the perception stage a depth-wise 3x3 convolution with a set of fixed (non-learned) kernels. The per-cell update ($f$) is a sequence of 1x1 convolutions with a ReLU. The additive update is often referred to as a "residual network", and even the stochastic discarding of updates for some cells can be thought of as a variant of dropout, applied per-cell, rather than per-value.

Once the image generator is expressed in terms of standard differentiable building blocks, we can use back-propagation from a provided objective function to learn the model parameters.

\paragraph{Computational efficiency}
NCA models described here are relatively compact by the modern standards. They contain less that 10k trainable parameters. We also use the quantization-aware training \cite{Jacob2018QuantizationAT} to make sure that our models can be efficiently executed on the hardware that stores both  parameters and activations as 8-bit integers. This allowed us to develop a WebGL-demo that allows to interact with learned NCA models in real time. We refer readers to the supplemental materials and the code release\footnote{\label{note1}\url{https://selforglive.github.io/cvpr_textures/}}.

\paragraph{Parameters}
The cell-state vector size (including visible RGB) is $\mathbf{s_t \in \mathbb{R}^{12}}$. Perception vector size is $4*12$; $\mathbf{p \in \mathbb{R}^{48}}$. The hidden layer size is \textbf{96}. Thus, matrices $W_0, W_1$ have dimensions \textbf{48x96} and \textbf{96x12}. Total number of CA parameters is \textbf{5868}.

\section{Training the Neural CA}

\subsection{Objective}
In order to train a NCA we need to define differentiable objective (loss) functions, that measure the current performance of the system, and provide a useful gradient to improve it. We experiment with two objectives - a VGG-based texture synthesis loss \cite{Gatys2015TextureSU} and an Inception-based feature visualisation loss \cite{olah2017feature}. Hereinafter, we refer to these as "texture-loss" and "inception-loss", respectively. These losses are applied to the snapshots of CA grid state $\mathbf{s}$, and are only affected by the first three values of state vectors, that are treated as RGB-color channels.

\paragraph{Texture Loss} Style transfer is an extensively studied application of deep neural networks. L. Gatys et al.  \cite{Gatys2015TextureSU} introduced the approach common to almost all work since - recording and matching neuronal activations in certain layers of an "observer" network - a network trained to complete a different task entirely whose internal representations are believed to capture or represent a concept or style. We apply the same approach to training our NCA. We initialize the NCA states as vectors of uniform random noise, iterate for a stochastic number of steps and feed the resulting RGB channels of the state into the observer network (VGG-16 \cite{Simonyan2015VeryDC}), and enforce a loss to match the values of the gram matrices when the observer network was fed the target texture and when it was fed the output of the NCA. We backpropagate this loss to the parameters of the NCA, using a standard backpropagation-through-time implementation \cite{bpp}.

\begin{figure}[th]
\begin{center}
  \includegraphics[width=0.4\linewidth]{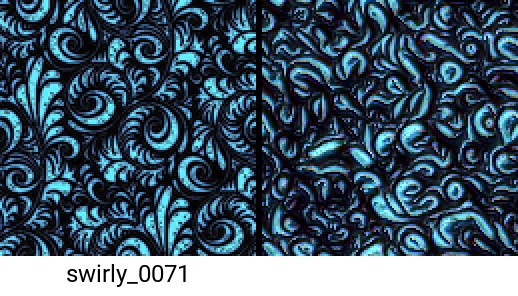}
  \includegraphics[width=0.4\linewidth]{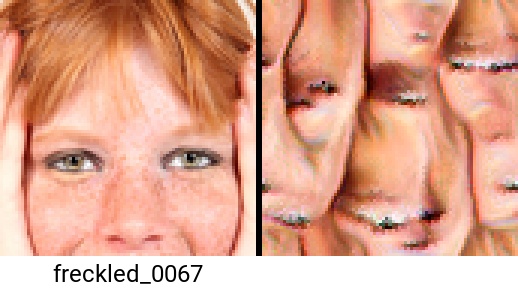}
  \includegraphics[width=0.4\linewidth]{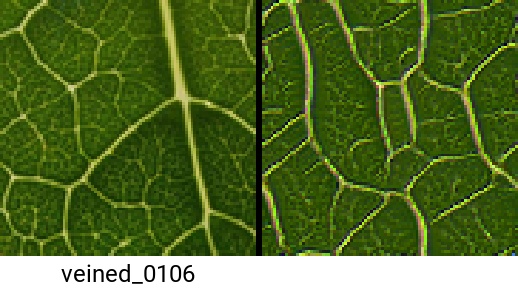}
  \includegraphics[width=0.4\linewidth]{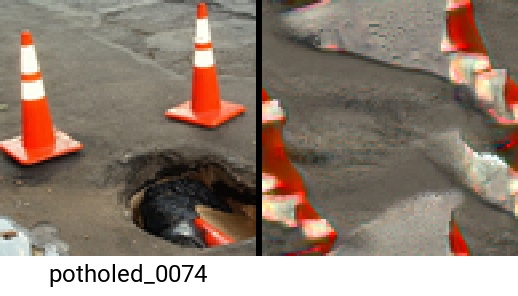}
  \includegraphics[width=0.4\linewidth]{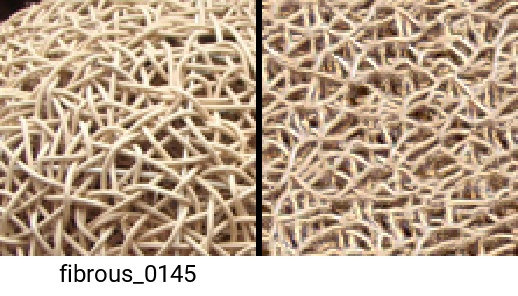}
  \includegraphics[width=0.4\linewidth]{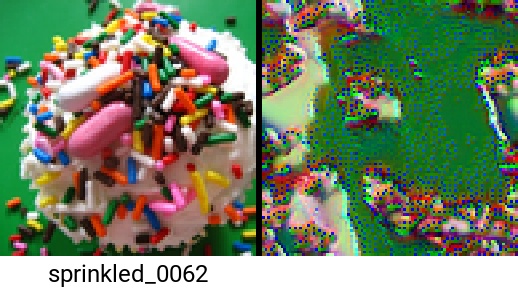}
 
 \end{center}
\caption{Some examples chosen to show good (left) and bad (right) convergence of the NCA. When the underlying template image is not consistent nor produced by a locally interacting system and instead requires global coordination, the NCA often fails.}
\label{fig:dtd_samples}
\end{figure}

\paragraph{Dataset} We use the textures collected in the Describable Textures Dataset by Cimpoi et al \cite{cimpoi14describing}. DTD has a human-annotated set of images relating to 47 distinct words describing textures, which in turn were chosen to approximate the high level categories humans use to classify textures. Each image has a primary descriptor label, as well as secondary and tertiary descriptors. We do not explicitly make use of the texture labels in this work, but we notice significant differences in the quality of the reproduction across the different texture categories. See \ref{fig:dtd_samples} for some examples of categories where our method fails to produce a coherent output. We note that these tend to be images representing textures that aren't the result of locally interacting processes, such as human faces or singular images of large potholes.

\paragraph{Inception Loss} Deepdream \cite{Mordvintsev2015InceptionismGD} and subsequent works have allowed insight into the features learned by networks, in addition to opening the doors to an extensive set of artistic and creative works \cite{mordvintsev2018differentiable}. We investigate the behaviours learned by NCA when tasked with maximizing certain neurons in an observer network. We use Inception \cite{Szegedy2015GoingDW} as an observer network and investigate the resulting behaviours for a variety of layers in this network. In the results section we show some of the more remarkable patterns generated using this loss.

\subsection{Training procedure}

\begin{figure}[t]
\begin{center}
   \includegraphics[width=0.95\linewidth]{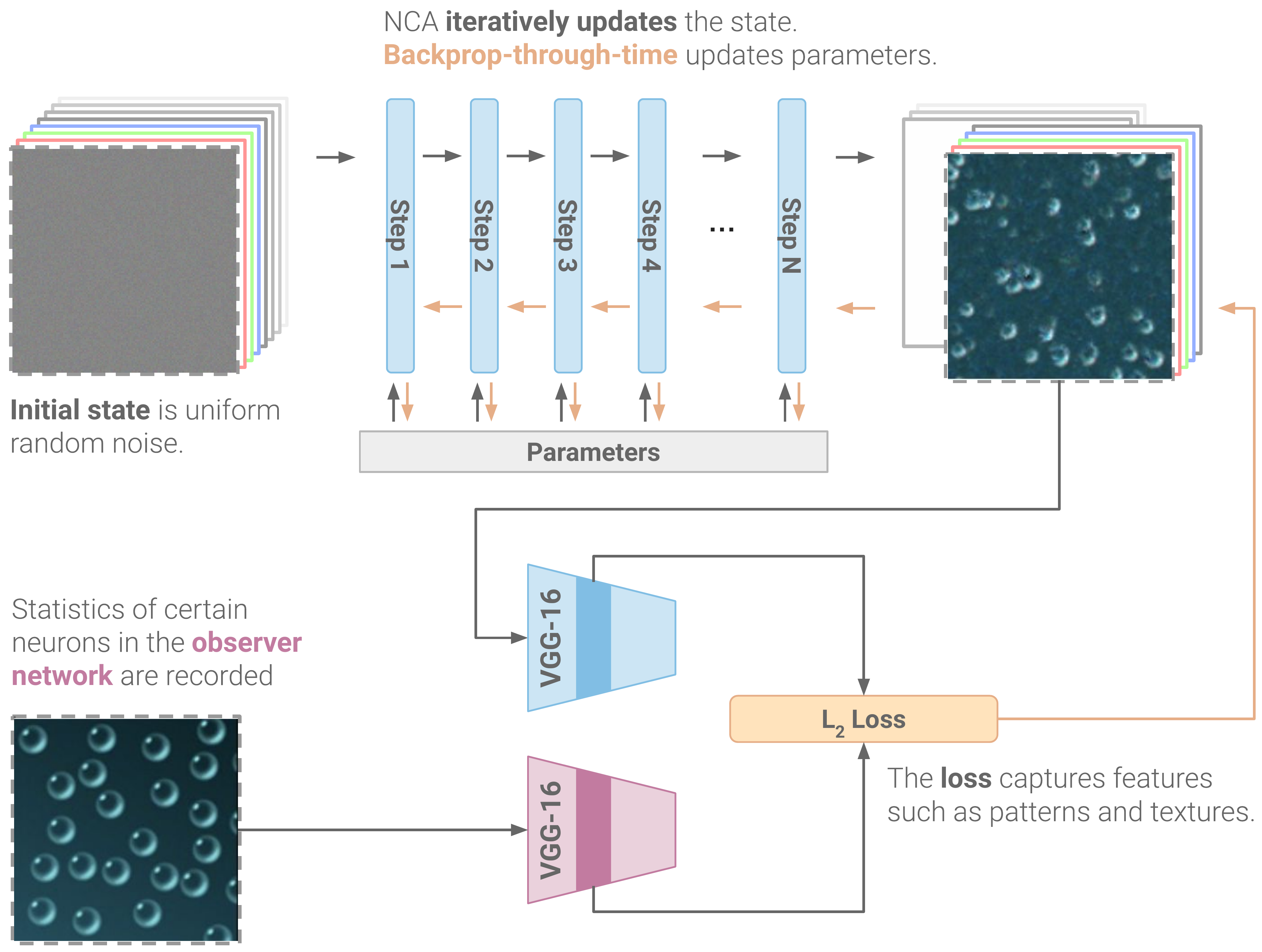}
\end{center}
\caption{Texture NCA training iteration with VGG-based texture loss.}
\label{fig:model_e2e}
\end{figure}

At each training iteration we sample a batch of initial grid states $\mathbf{s_0}$, and iterate the current NCA for $N \overset{\text{i.i.d.}}{\sim} \mathcal{U}\{32, 64\}$ steps. The batch loss is computed at $\mathbf{s}_N$ and backpropagation through time is used to adjust the CA parameters. Batches of \textbf{4} images, \textbf{128x128}, are used during training. The state checkpoint pool size is \textbf{1024}. The NCA model for each pattern is trained for \textbf{8000} steps using the Adam optimizer. Learning rate is \textbf{2e-3} and decays to \textbf{2e-4} after \textbf{2000} steps. A single texture CA trains in a few minutes on a V100 GPU.

\paragraph{State checkpointing} We'd like to ensure that the successive application of the learned CA rule doesn't destroy the constructed pattern over numbers of steps that largely exceed that of training runs. We adopt the checkpointing strategy from \cite{mordvintsev2020growing}. It consists of maintaining a pool of grid states, initialised with empty states. At each training step we sample a few states from the pool and replace one of them with an empty state, so the the model doesn't forget how to build the pattern from scratch. The final states are placed back into the pool, replacing the sampled ones.

An interesting feature of the proposed texture generation method is the lack of an explicitly defined final state of the process, nor the requirement to generate a static pattern after a series of CA steps. We only require individual snapshots to look like coherent textures. In practice, we observe that this leads to the emergence of "living", constantly evolving textures. We hypothesize that the NCA finds a solution where the state at each step is aligned \cite{mordvintsev2018differentiable} with the previous step and thus the motion, or appearance of motion, we see in the NCA is the state space traversing this manifold of locally aligned solutions.

\section{Results and discussion}

\subsection{Qualitative texture samples}

Below we present some of the learned textures which demonstrate creative and unexpected behaviours in the NCA. \textbf{While these snapshots provide some insight into the time-evolution of the NCA, we strongly urge readers to view the videos and interactive demonstrations in the supplementary materials.\addtocounter{footnote}{-1}\addtocounter{Hfootnote}{-1}\footnotemark}


\paragraph{Bubbles}

\begin{figure}[th]
\begin{center}
   \includegraphics[width=1.0\linewidth]{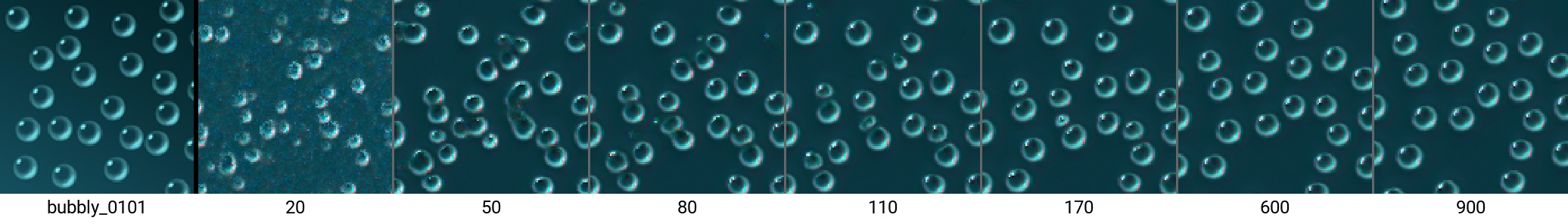}
 \end{center}
\caption{Bubbles are initially generated by the NCA as several small seeds of bubbles. Some of these grow to be larger bubbles, and some implode. Eventually, a bubble-density similar to the template image is achieved. The NCA exhibits behaviour where bubbles appear to behave almost like solid structures, despite being a visual product of a homogeneous computation on a grid which does not differentiate between background and bubble.}
\label{fig:bubbles_sample}
\end{figure}

Figure \ref{fig:bubbles_sample} shows the time-evolution of a texture-generating algorithm trained on the static image of several bubbles on a plain blue background. The resulting NCA generates a set of bubbles, moving in arbitrary directions, with the spatial density of bubbles in the image roughly corresponding to that of the static template image. It is important to bear in mind that the NCA knows nothing of the concept of bubbles existing as individual objects that can have their own velocity and direction. However the NCA treats them as such, not allowing them to collide or intersect. We refer to such structures as \textbf{solitons} in the solution space of NCA, named after the concept introduced to describe the structures and organisms found in the solution space of Lenia \cite{chan2019lenia} \footnote{Classically, solitons refer to self-reinforcing wave packets found in the solutions to certain PDEs. Chan borrowed this terminology for organisms in Lenia, and we borrow it for structures in our NCA.}.

\begin{figure}[th]
\begin{center}
   \includegraphics[width=1.0\linewidth]{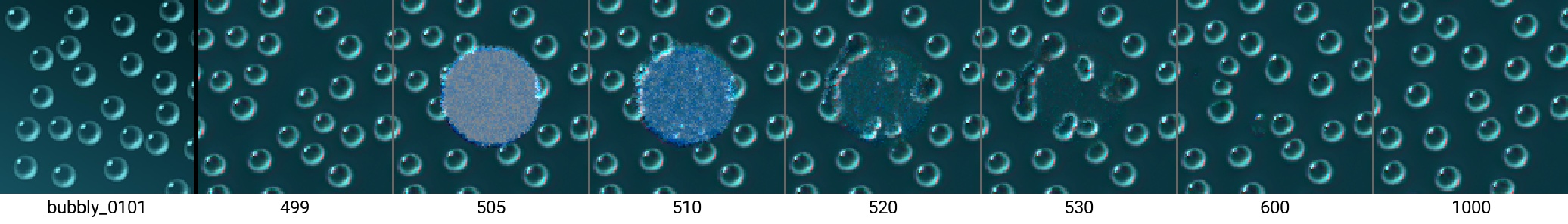}
 \end{center}
\caption{A converged bubble NCA has a circular section of the state manifold set to uniform random noise. The NCA recovers and recreates bubbles in this section.}
\label{fig:bubble_break}
\end{figure}

Figure \ref{fig:bubble_break} shows the behavior of the NCA when some bubbles are destroyed by setting the states of these cells to random noise. The rest of the pattern remains unchanged and stable, and over the course of a few time steps, a somewhat consistent pattern is filled in inside the gap. Many new bubbles appear and some are destroyed when there is crowding of the bubbles, eventually returning to a "bubble-density" that roughly corresponds to that of template image. Some of the new bubbles immediately after the damage are misshapen - oblong or almost divided into two. Misshapen bubbles have the ability to recover their shape, or in severe cases divide into smaller bubbles. 

\paragraph{Grids}

\begin{figure}[th]
\begin{center}
   \includegraphics[width=1.0\linewidth]{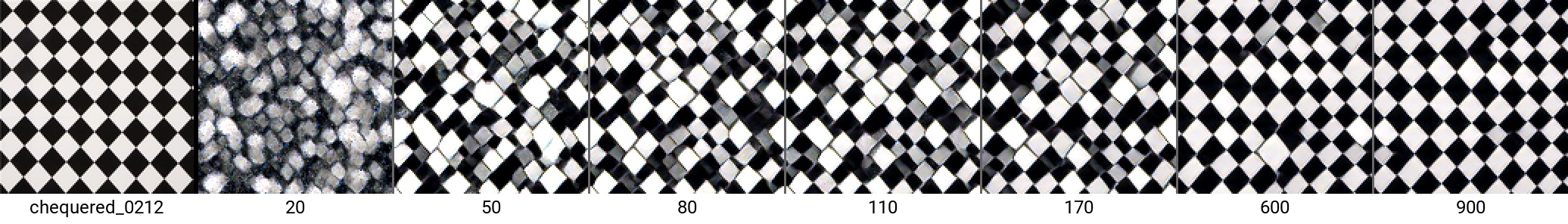}
 \end{center}
\caption{First, candidate black and white diamonds are generated. These are then iteratively merged or destroyed until perfect consistency is achieved.}
\label{fig:grid_diag}
\end{figure}

Figure \ref{fig:grid_diag} shows behaviour trained on a chequered diamond pattern. The NCA generates a number of potential black and white diamonds. At first, these are randomly positioned and not aligned, resulting in a completely inconsistent grid. After a few iterations, certain diamonds are removed or joined with each other, resulting in more consistency. After a long time period ($N_{evaluation} \gg N_{training}$), in this case approximately $1e3$ steps, we see the grid reaching perfect consistency, suggesting the NCA has learned a distributed algorithm to continually strive for consistency in the grid, regardless of current state.

\paragraph{Mazes}

\begin{figure}[th]
\begin{center}
   \includegraphics[width=1.0\linewidth]{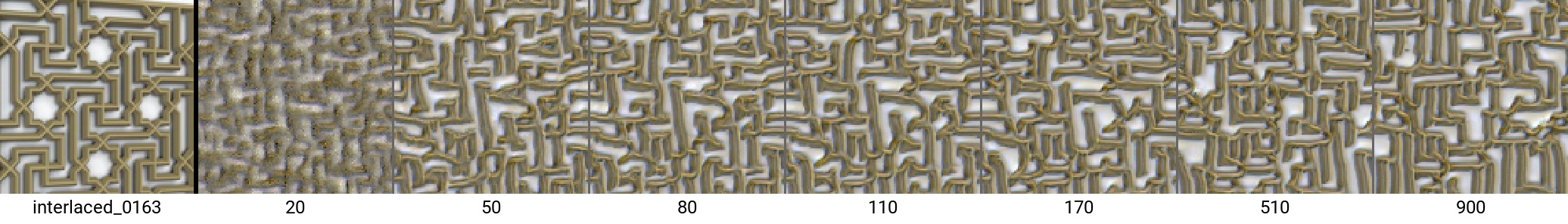}
 \end{center}
\caption{The walls of this maze-like texture are treated as solids. They migrate and join other walls to form the texture, incorporating some of the path-morphology of the original walls.}
\label{fig:mazelike}
\end{figure}

In figure \ref{fig:mazelike}, the NCA learns the core concept of a wall, or barrier, as a soliton. A similar behaviour can be observed on the 6th row of the front-page figure. The walls proceed to move in arbitrary directions, and when a free edge reaches another wall, it joins to form a vertex. The resulting pattern appears random, but incorporates many of the details of the template image, and has successfully learned the "rules" of walls in the input texture (they remain of a fixed width, they tend to be aligned in certain directions, and so forth). 

\paragraph{Stripes}

\begin{figure}[th]
\begin{center}
   \includegraphics[width=1.0\linewidth]{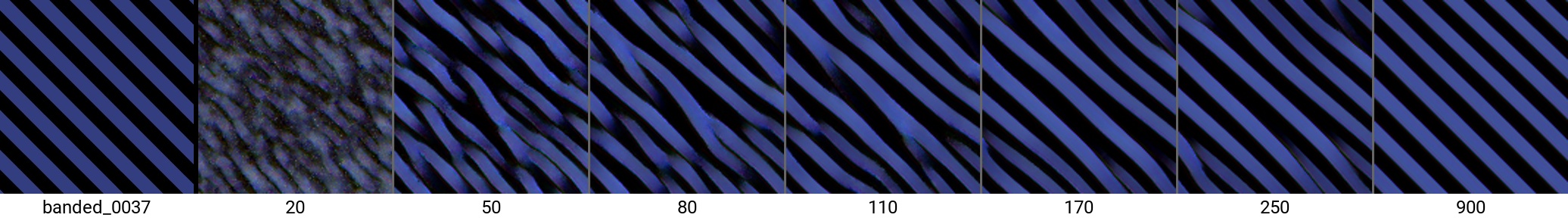}
 \end{center}
\caption{The NCA starts by forming multiple potential blue stripes, which subsequently merge with one other until total consistency is reached.}
\label{fig:stripes}
\end{figure}

In figure \ref{fig:stripes}, a distributed algorithm emerges which tries to merge different stripes to achieve consistency. Ill-fitting stripes with free ends travel up or down along the diagonal, until they either spontaneously merge with the neighbouring stripe, or find another loose end to merge with. Eventually, total consistency is achieved. 

\paragraph{Triangular mesh}

\begin{figure}[th]
\begin{center}
   \includegraphics[width=1.0\linewidth]{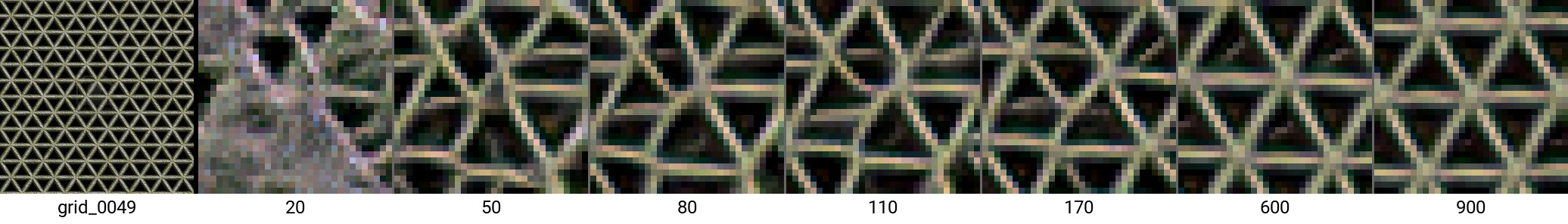}
 \end{center}
\caption{8x zoomed view of NCA exhibiting algorithmic behaviour over time - rearranging strokes to achieve perfect consistency in the lattice.}
\label{fig:triangles}
\end{figure}

Figure \ref{fig:triangles} depicts the generation of a triangular mesh. As with other templates, at first the NCA generates a multitude of candidate strokes in the pattern. These strokes then either traverse along each other, disconnect, or rejoin other vertices in order to increase consistency in the pattern. Their motion appears smooth and organic in nature. After a longer time, the generated texture approaches perfect consistency. 

\paragraph{Weave}

\begin{figure}[th]
\begin{center}
   \includegraphics[width=1.0\linewidth]{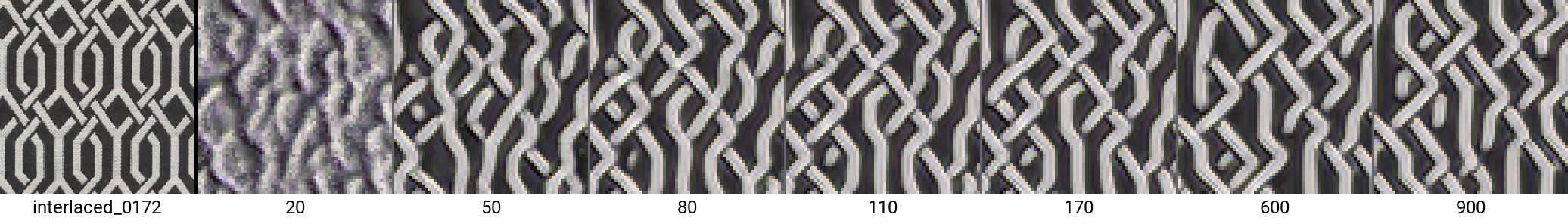}
 \end{center}
\caption{The NCA forms a complicated threading pattern by first generating curved threads, snapping them to the diagonal and vertical directions, then joining them either "over" or "under" other threads.}
\label{fig:viking}
\end{figure}

In figure \ref{fig:viking}, the NCA attempts to learn to generate a pattern that consists of a weave of different thread crossing each other. The NCA captures this rule of threads being oriented in one of three directions - the diagonals or the vertical, and generates a texture similar in style to the original. However, it does not exactly capture the repeating pattern of the template texture.

\subsection{Inception samples}

\begin{figure}[th]
\begin{center}
   \includegraphics[width=1.0\linewidth]{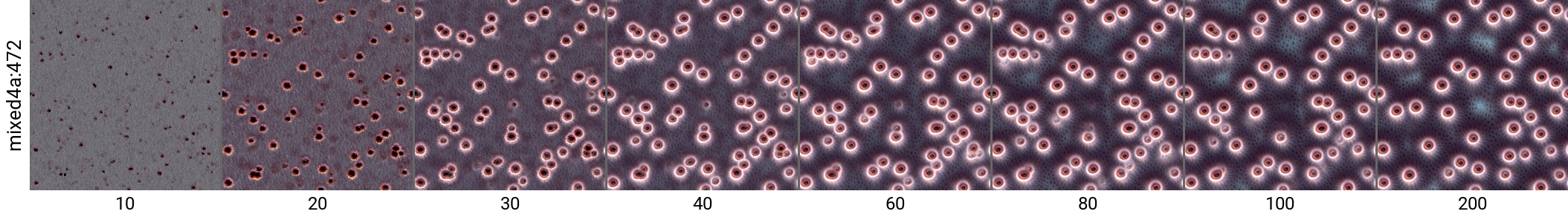}
    \includegraphics[width=1.0\linewidth]{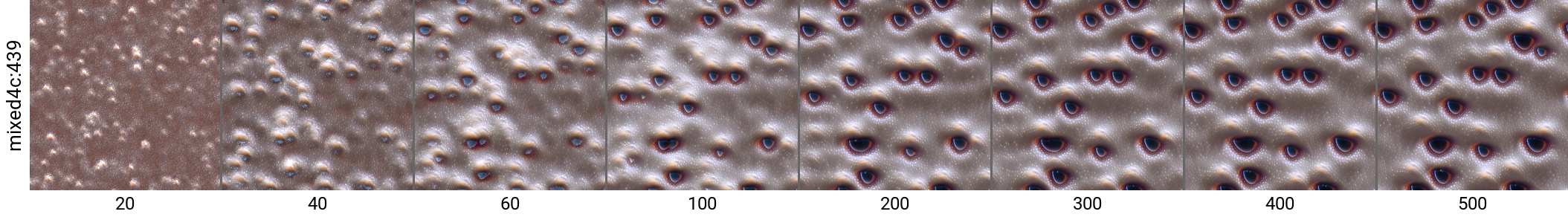}
    \includegraphics[width=1.0\linewidth]{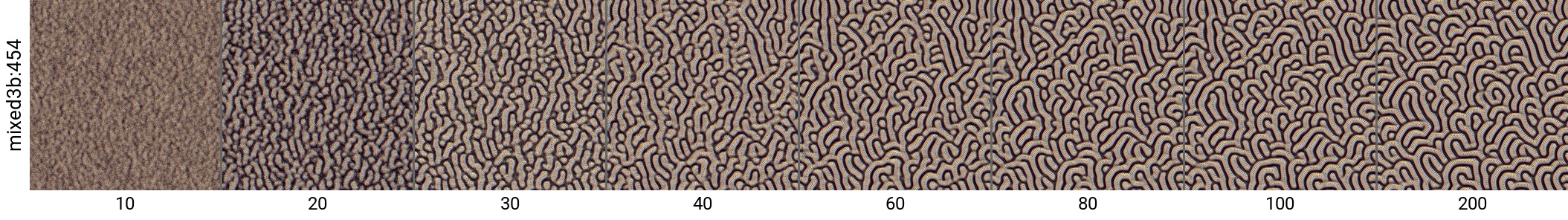}
    \includegraphics[width=1.0\linewidth]{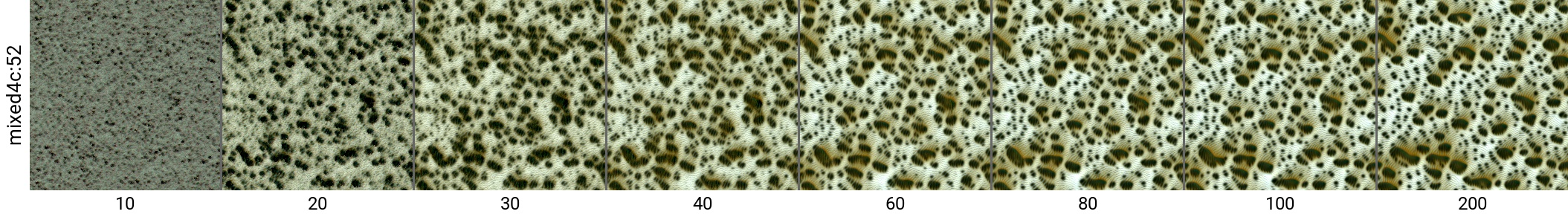}
    \includegraphics[width=1.0\linewidth]{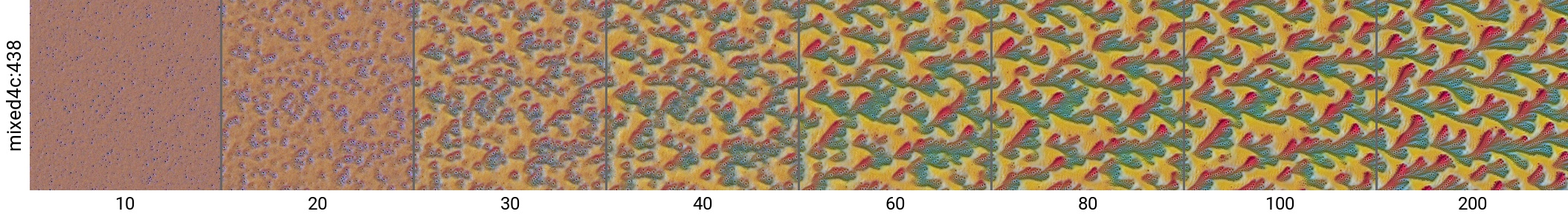}
    \includegraphics[width=0.15\linewidth]{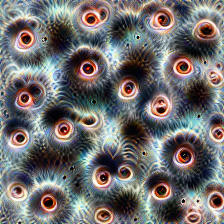}
    \includegraphics[width=0.15\linewidth]{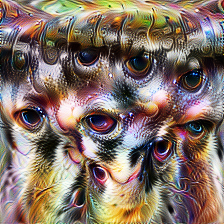}    
    \includegraphics[width=0.15\linewidth]{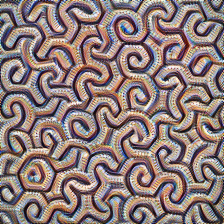}    
    \includegraphics[width=0.15\linewidth]{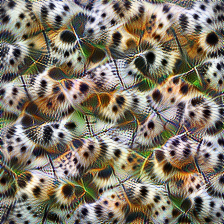}    
    \includegraphics[width=0.15\linewidth]{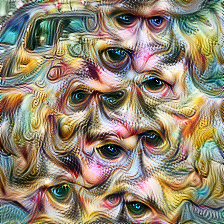}    
 \end{center}
\caption{NCA trained with the inception feature visualisation loss, exhibiting a variety of behaviours, often forming different solitons, such as eye-like shapes, and producing consistent patterns, such as interlaced "corals". In the bottom row are visualizations of the same patterns in order, from OpenAI Microscope \cite{cammarata2020thread}. Some NCA results demonstrate strong similarity with their image-space optimized counterparts, while others are very different.}
\label{fig:inception}
\end{figure}

Figure \ref{fig:inception} shows a small selection of patterns, obtained by maximising individual feature channels \cite{olah2017feature} of the pre-trained Inception network. Some of these patterns exhibit similar behaviour to those trained with the texture-loss - merging or re-arranging different solitons to achieve some form of consistency in the image. 

\subsection{Advantages of NCA for texture generation}
\paragraph{Distributed algorithms for texture generation}

Neural GPUs and Neural Turing Machines introduced to a wider audience the idea of deep neural networks being able to learn algorithms, as opposed to just serving as excellent function approximators. We believe this is an underappreciated method shift that will serve to allow deep learning solutions to solve more complex problems than is currently possible, due to the more general nature of an algorithm as opposed to an approximated function. Similar observations have been made about RNNs \cite{rnns}, but empirical results in terms of learning computations versus statistical correlations have been weak. 

We believe an NCA inherently learns a \textbf{distributed algorithm} to solve a task. Any single cell can only communicate with neighbouring cells, a key first skill the cells must learn is an algorithm for coordination and information sharing over a longer period of time. The stochastic updates encourage any such algorithm to be resilient and inherently distributed in nature - it must perform its task regardless of when or where the cell may be invoked.

The results section present several qualitative observations of behaviour we think shows evidence of a distributed algorithm having been learned.

\paragraph{Long term stability}

Recall the model and training regime exposes the NCA to a loss after $n \in \overset{\text{i.i.d.}}{\sim} \mathcal{U}\{32, 64\}$ steps. Longer time period stability is encouraged by the sample pool mechanism. However, we observe the solutions learned by the NCA to be stable for far longer time periods than those used during training, even accounting for the sample pooling mechanism. This suggests the learned behaviour enters a stable state where minor deviations return back to the same solution state (a basin of attraction \cite{mordvintsev2020growing}).

\begin{figure}[t]
\begin{center}
   \includegraphics[width=0.95\linewidth]{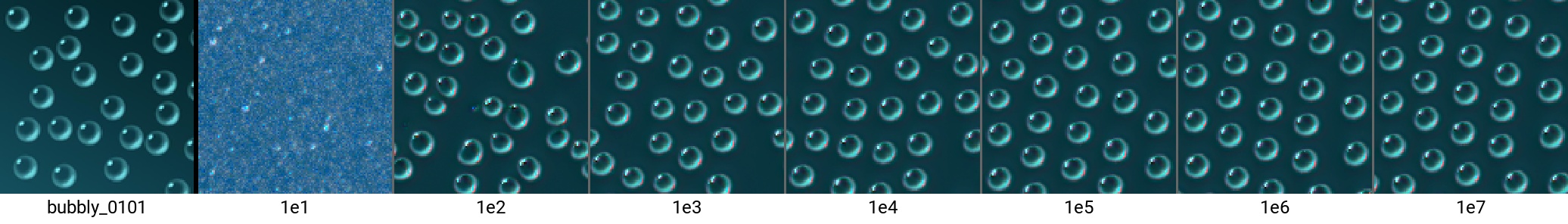}
\end{center}
\caption{NCA evaluated up to $1e7$  steps, showing stability of the learned texture. Notice that the non-stationary behaviour slows down but doesn't stop - the bubbles are always ever-so-slightly in motion.}
\label{fig:longterm}
\end{figure}

Figure \ref{fig:longterm} shows an NCA evaluated for $1e5$, $1e6$ and $1e7$ steps. We believe most NCA we have trained on textures are fully stable over longer time periods on this time-scale.

\paragraph{Spatial invariance}

\begin{figure}[t]
\begin{center}
   \includegraphics[width=0.95\linewidth]{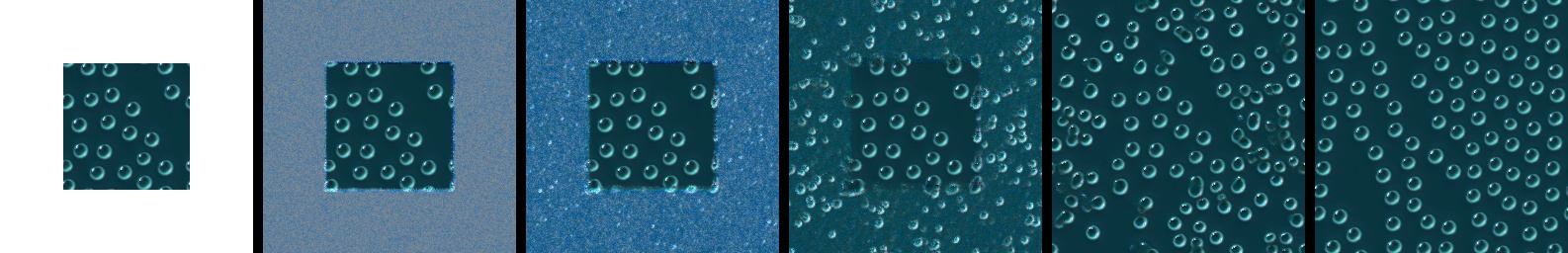}
   \includegraphics[width=0.95\linewidth]{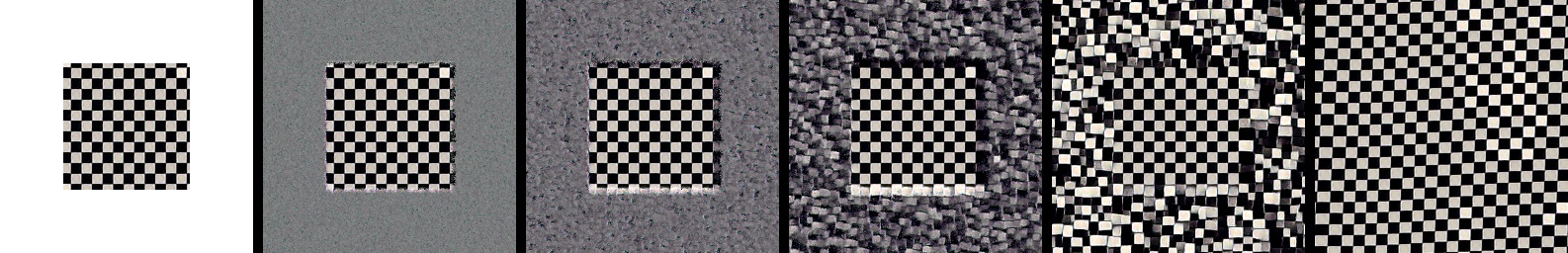}
\end{center}
\caption{NCA generated texture on a 128x128 grid, which is then expanded to 256x256. The NCA fill in the missing texture and proceeds to finds consistency with the existing cells.}
\label{fig:expand}
\end{figure}

NCA are spatially invariant - it is possible to extend them in any direction with linear complexity on the order of number of pixels and number of time-steps. The computation can continue even with irregular or dynamic spatial boundaries. Figure \ref{fig:expand} shows a fixed size NCA being expanded to double the width and height. The newly added cells are initialised in the usual way, with uniform random noise, and run the same NCA rule as the existing cells. The NCA immediately fills out the empty space, interacting with the existing cells to form a continuation of the pattern consistent with the initial texture (i.e. the newly formed checkerboard spaces align themselves with the existing grid). 

The NCA is further employable as a "texture printer" - generating the first $N$ rows of an output pattern, "freezing" these cells by stopping them from undergoing any further updates, then evaluating the next rows of cells. The next rows of cells would only rely on the $N$:th cell for the information necessary to continue generating the pattern in a consistent fashion. Thus, arbitrarily large, consistent, textures can be generated with only a linear computational cost and a linear memory cost to store the "finished" cells, without the need for the entire grid to be computed at once to achieve consistency.

\paragraph{Parallelism}

\begin{figure}[t]
\begin{center}
   \includegraphics[width=0.95\linewidth]{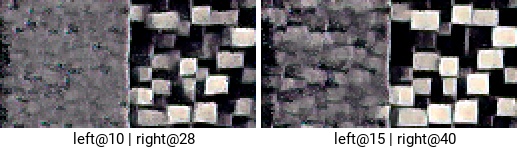} \\
   \includegraphics[width=0.95\linewidth]{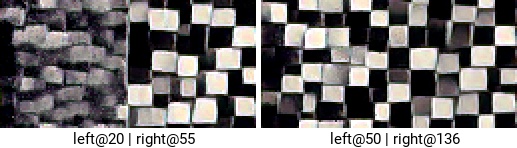} \\
    \includegraphics[width=0.95\linewidth]{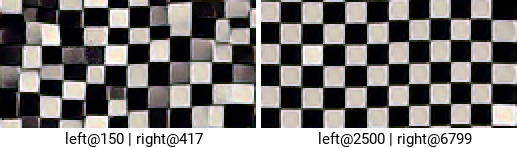}
\end{center}
\caption{Two instances of an NCA running in parallel, at different wall-clock speeds. Each operate on the left and right half of the space, respectively. The left NCA and right NCA can also access the left-most, and right-most column of cells, respectively of the opposite NCA. The right NCA is stochastically evaluated at a rate approximately $e$ times the rate of the left NCA. The label denotes the current time step of each NCA at the time of the snapshot. Despite running asynchronously, the NCA effectively communicates between the two instances using these two columns of mutually accessible cells, and forms a completely consistent pattern.}
\label{fig:parallel}
\end{figure}

NCA are evaluated in a highly parallel fashion. For instance, one could have two NCA running in parallel on separate hardware coordinate spatially by simply sharing the boundary layer of cells between them. Synchronisation in time is not required as the NCA are robust to asynchronous updates, as can be seen in  figure \ref{fig:parallel}.

\paragraph{Robust to unreliable computations}
Thirdly, the resulting algorithm is extremely robust. For instance, it is possible to delete individual cells or groups of cells, or add individual cells at the boundaries. We demonstrate this behaviour in in the Results section in figure \ref{fig:bubble_break} as well as on several NCA in the first-page figure. NCA are thus ideal for any unreliable underlying computational hardware - many cells can fail or be reset, and they will "heal" in a fashion that is consistent with the existing pattern. Section \ref{section:post_training} further explores this property by altering the grid underlying the computation.

\subsection{Post-training behaviour control}

\label{section:post_training}
\begin{figure}
    \begin{center}
    \includegraphics[width=0.95\linewidth]{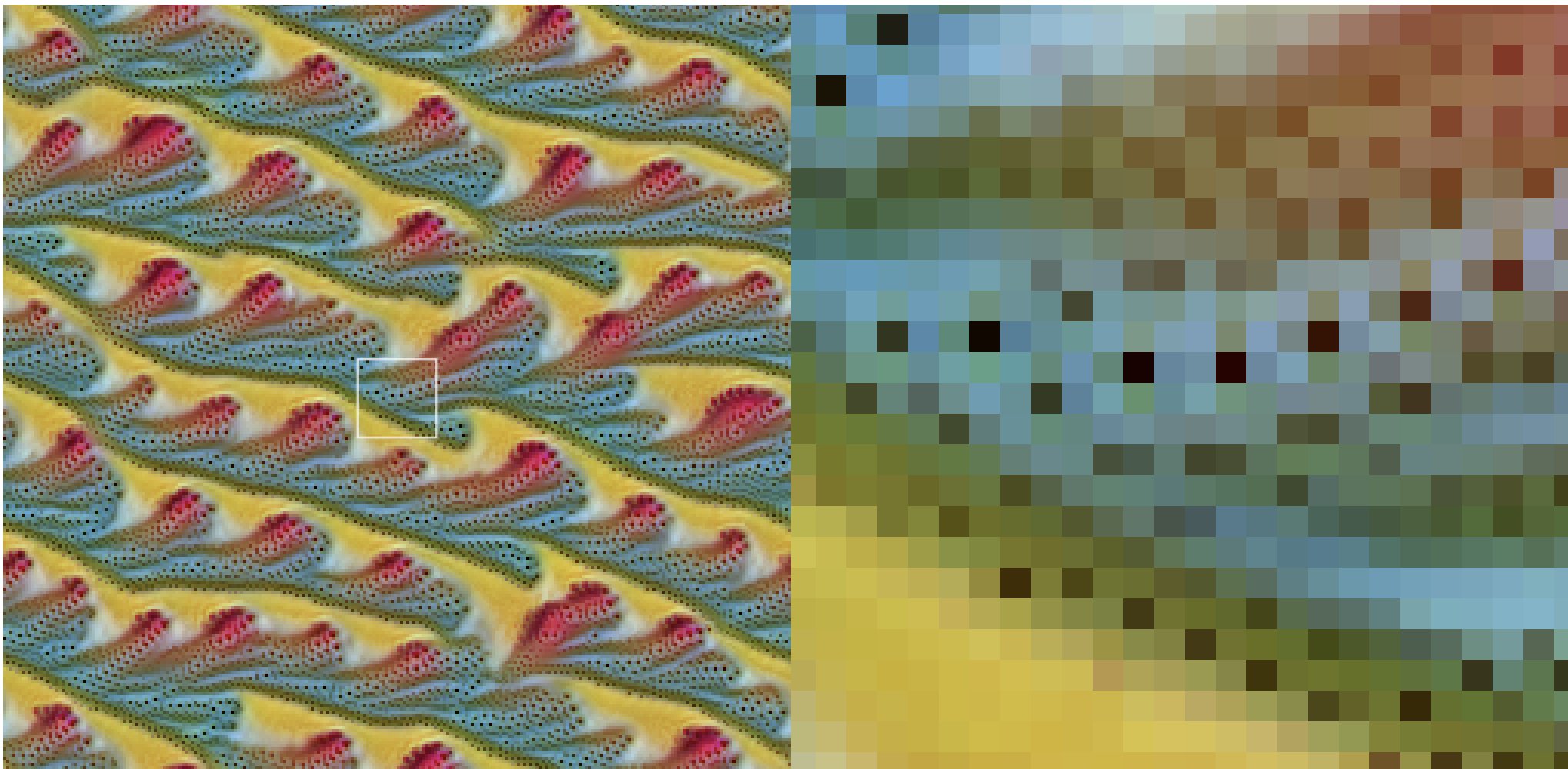}
    \includegraphics[width=0.95\linewidth]{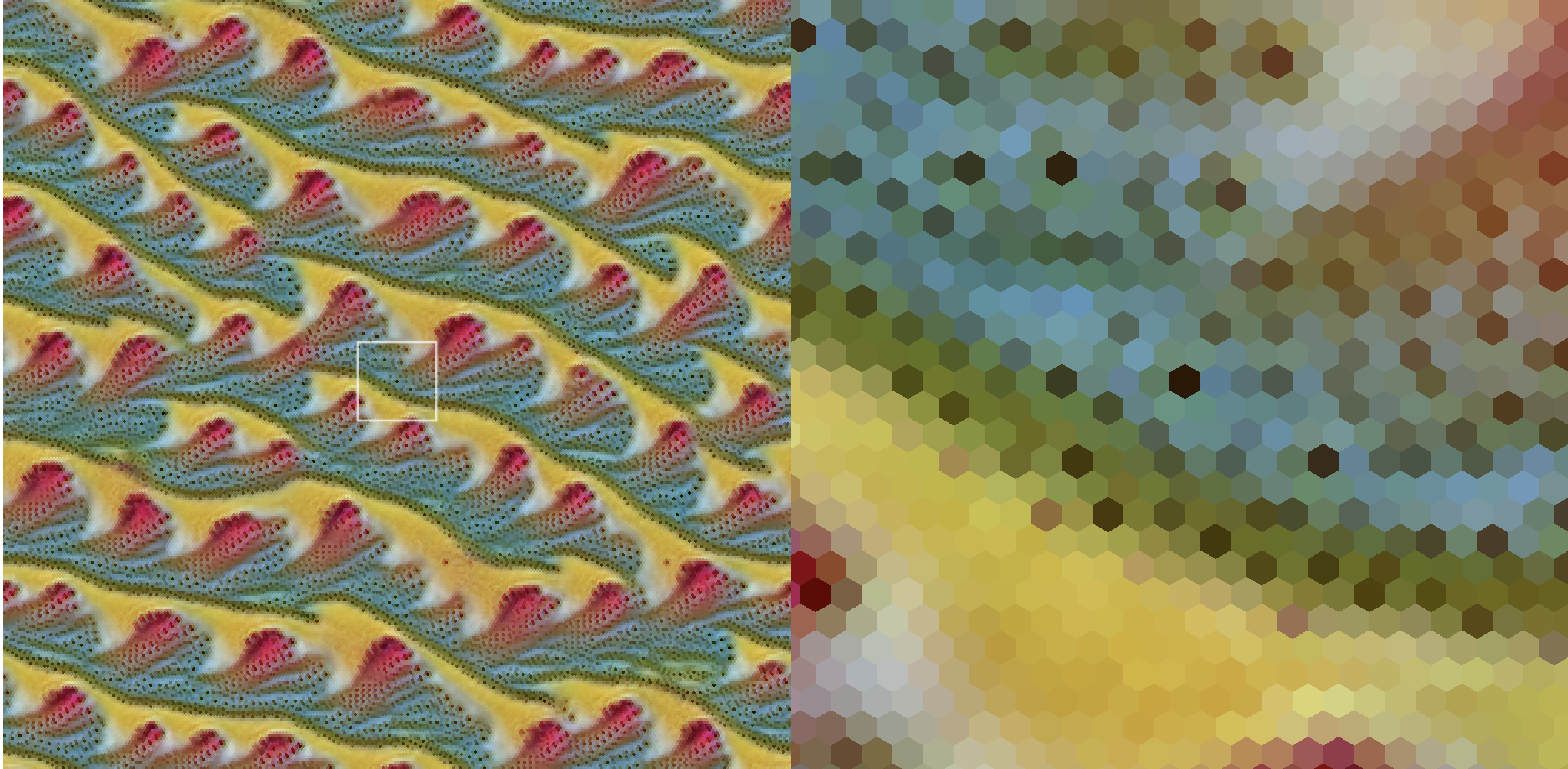}    
    \includegraphics[width=0.95\linewidth]{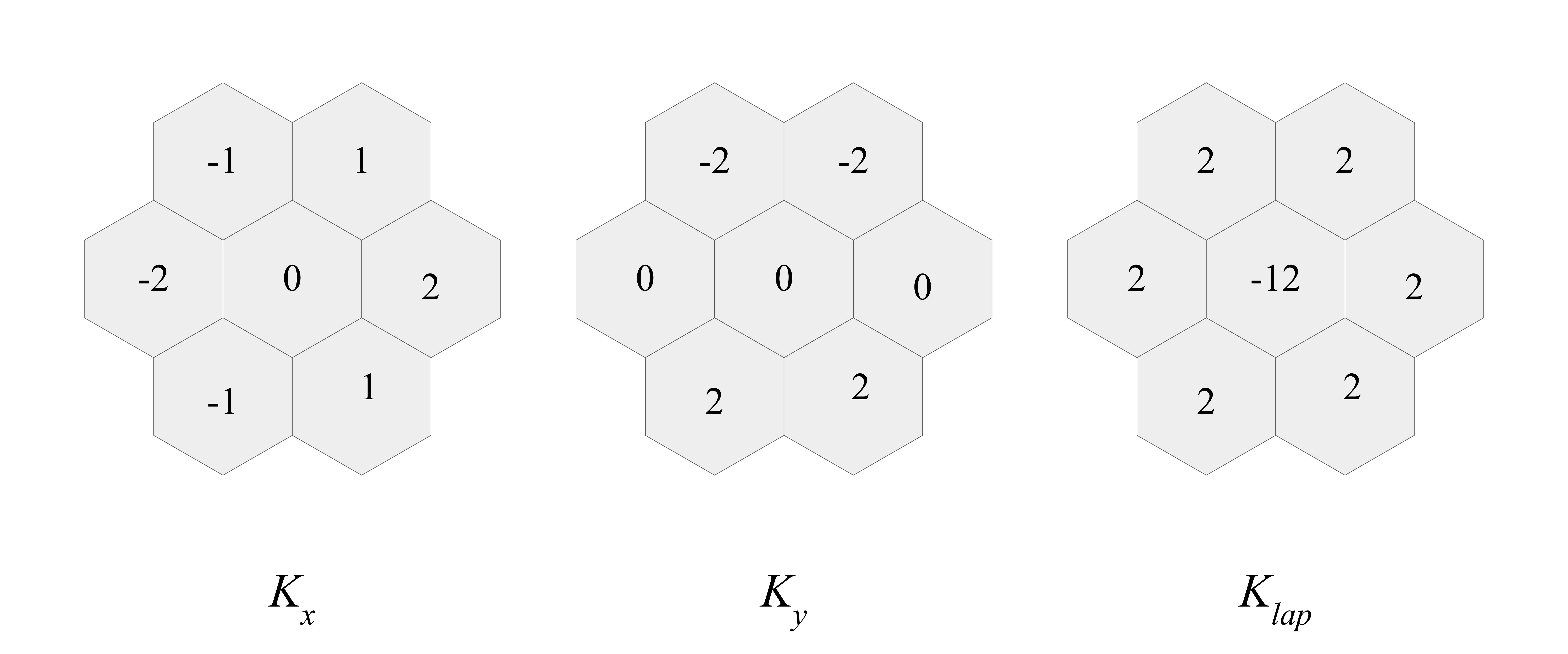}
    \end{center}
    \label{fig:hex_kernels}
    \caption{The same CA rule running on the square grid (top row) and the hexagonal grid (middle row). The bottom row shows the hexagonal equivalents of the perception convolution kernels.}
\end{figure}

As mentioned in the section \ref{section:ca_model}, trained CA cells obtain the information about their neighborhood through local gradient and Laplacian operators. This opens a number of opportunities for post training model adaptation and transfer to new environments. For example, we demonstrate the possibility of replacing the square grid with a hexagonal one just by using a different set of convolution kernels (fig. \ref{fig:hex_kernels}).

\begin{figure}
    \centering
    \includegraphics[width=0.48\linewidth]{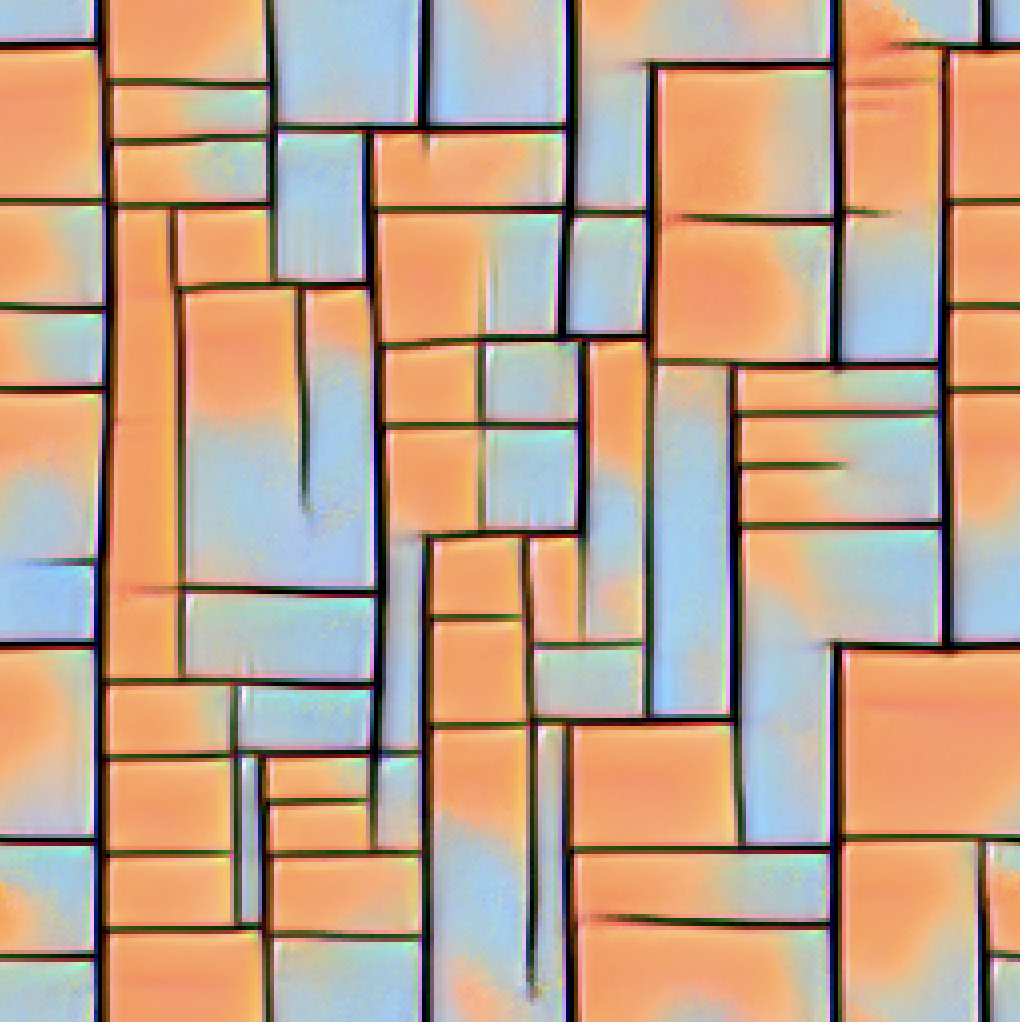}
    \includegraphics[width=0.48\linewidth]{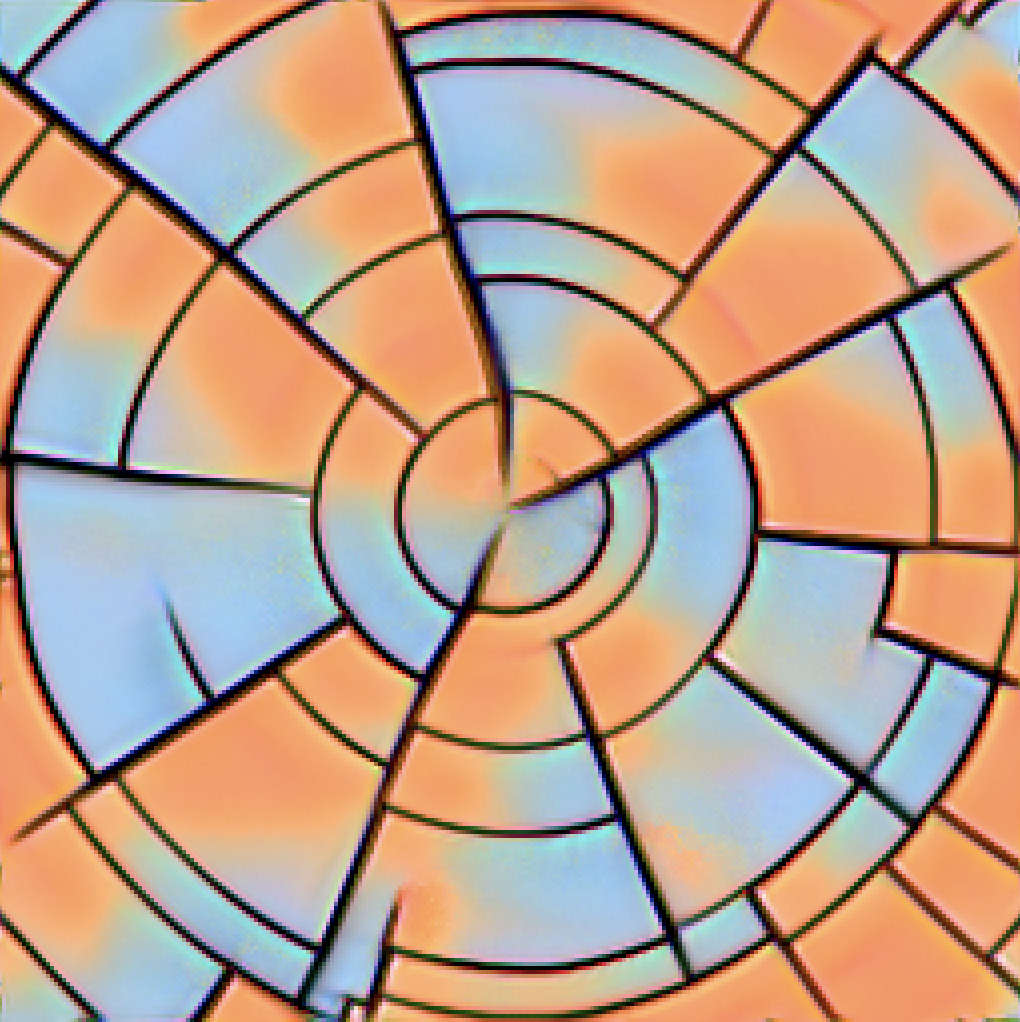}
    \includegraphics[width=0.48\linewidth]{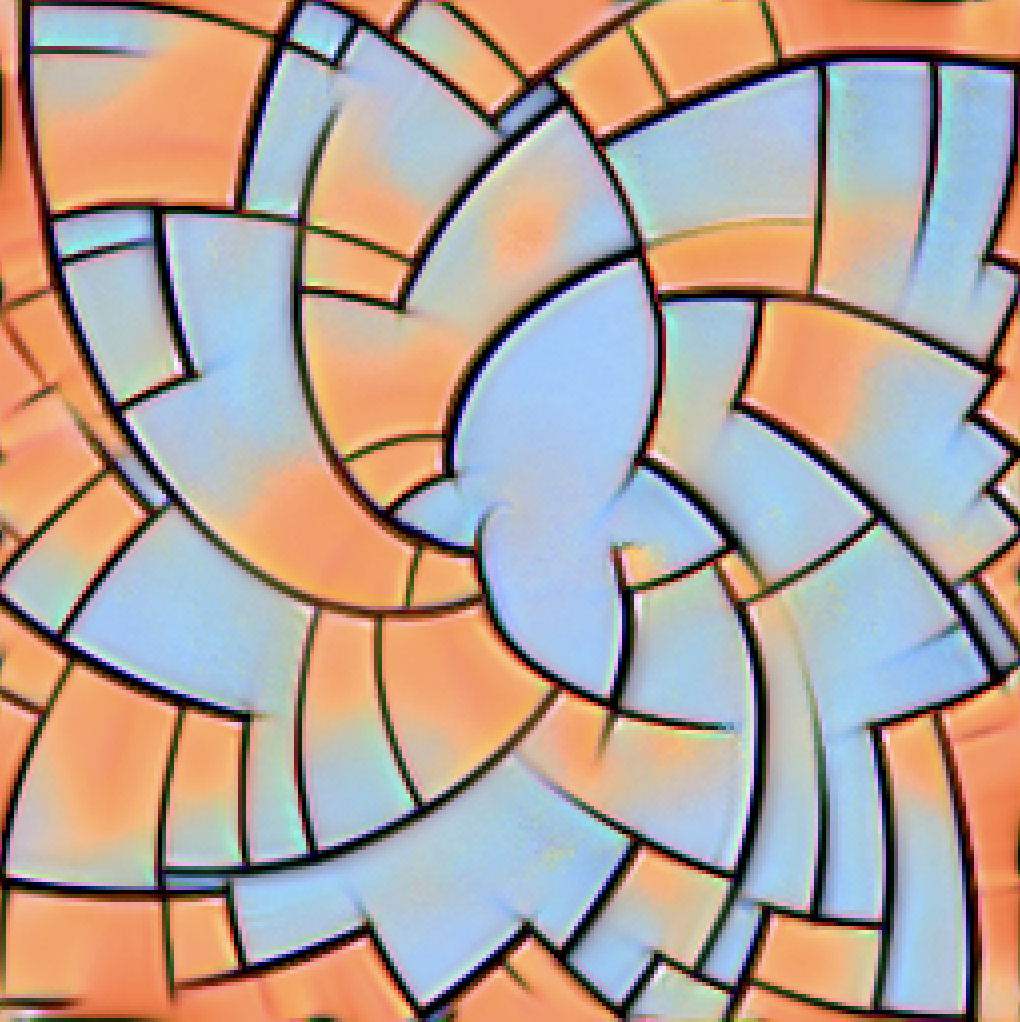}
    \includegraphics[width=0.48\linewidth]{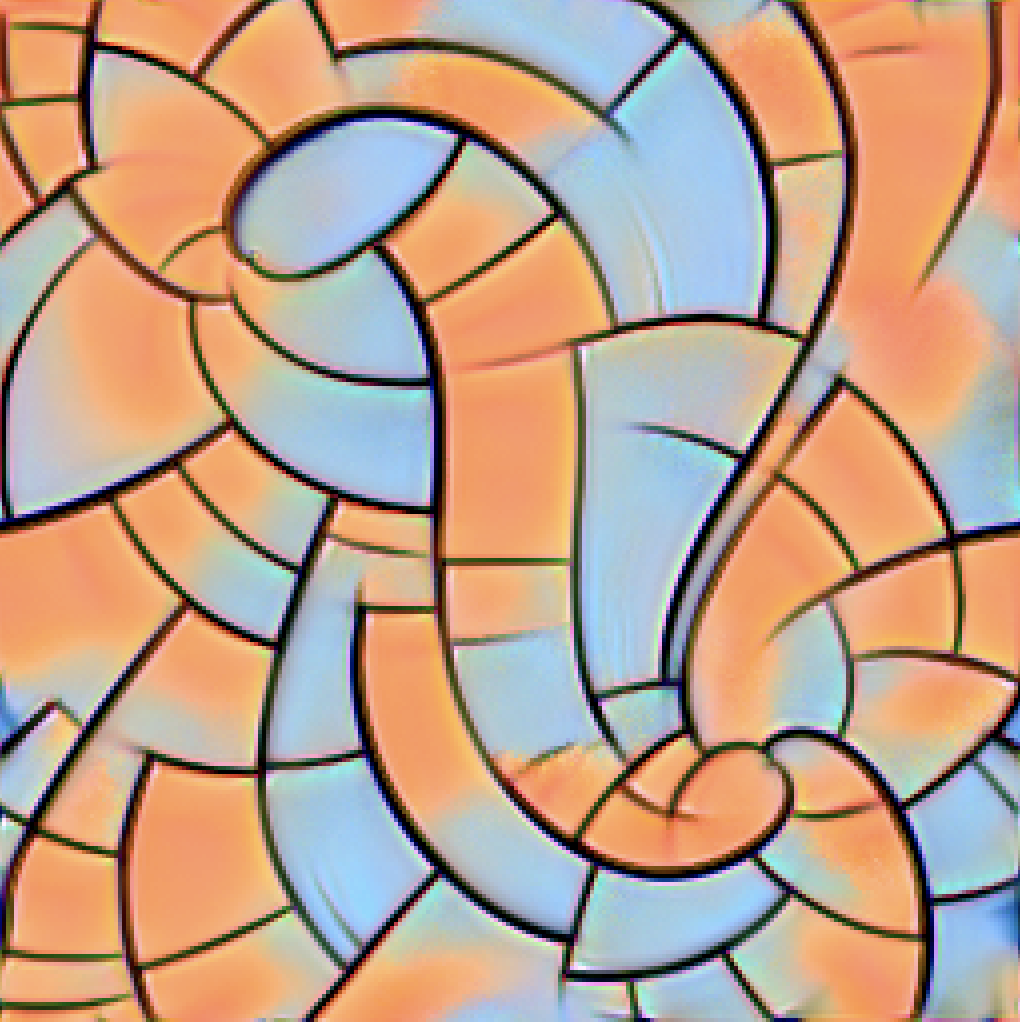}
    \caption{Patterns can be transformed with a vector field by applying per-cell rotations to the gradient vectors, estimated with the convolution kernels. Note that the NCA is not re-trained - it generalises to this new rotated paradigm without issue.}
    \label{fig:rotations}
\end{figure}

Another adaptation possibility, inspired by \cite{Lefebvre2006AppearancespaceTS}, is enabled by rotating the estimated local gradient vectors before applying the update rule:
$$
\begin{bmatrix}
\mathbf{g}_x \\ \mathbf{g}_y
\end{bmatrix}
=
\begin{bmatrix}
c & s \\ -s & c
\end{bmatrix}
\begin{bmatrix}
K_x \ast \mathbf{s} \\ K_y \ast \mathbf{s}
\end{bmatrix},
$$where $s=sin(\alpha_{x,y}), c=cos(\alpha_{x, y})$, and $\alpha_{x,y}$ is a local rotation angle for the cell at position $(x, y)$. This trick allows to project the texture onto an arbitrary vector field (fig. \ref{fig:rotations})

\section{Related work}



Probably the most popular family of methods for solving this task is based on sampling new pixels or patches from the provided example, conditioned on the already synthesized part of the output image \cite{Efros1999TextureSB,Gumin2016, Barnes_Shechtman_Finkelstein_Goldman}. Another family of methods is based on the iterative image optimisation process that tries to match output image feature statistics, produced by some extractor, to those of the input sample. The extractor can be hand-designed \cite{Portilla2004APT}, or based on a pretrained convolutional neural network \cite{Gatys2015TextureSU}. In either case, the output of texture synthesis is the raster image.

A very unconventional approach to texture synthesis is presented by C. Reynolds in \cite{Reynolds2011InteractiveEO}. Authors use human guided selection to steer the evolution process to create image patches that are hard to distinguish from the textured background. This work, along with a few others \cite{Sims1991ArtificialEF,Stanley2007CompositionalPP,ha2016abstract}, uses function composition instead of a raster grid as an image representation.

The idea of the importance of the right image representation for image processing tasks in the differentiable optimisation context is explored further in \cite{mordvintsev2018differentiable}. Another line of research focuses on employing image-generating neural networks to represent a whole family of texture images \cite{Ulyanov2016TextureNF}.

There is existing work attempting to learn rules for CAs that generate target images \cite{Elmenreich_Fehervari_2011}. Our method uses backpropagation instead of evolution,
and focuses on textures rather than pixel-perfect reconstructions. Authors of \cite{Elmenreich_Fehervari_2011} admit inability of their method to synthesize textures in the last paragraph of section five.

Other related work includes \cite{Henzler_Mitra_Ritschel_2019} and \cite{Yu_Barnes_Shechtman_Amirghodsi_Lukac_2019}; both make use of an encoder-decoder architecture to construct a texture generator. Also related is a family of work using GAN-style approaches for texture generation; \cite{Portenier_Bigdeli_Goksel_2020}, \cite{Bergmann_Jetchev_Vollgraf_2017}, \cite{Zhou_Zhu_Bai_Lischinski_Cohen-Or_Huang_2018}.

A very interesting image restoration method proposed in \cite{Chen2017TrainableNR}. The authors use a learned PDE system, that is spatially (but not temporally uniform).

\section{Conclusion}
Inspired by pattern formation in nature, this work applies an augmented version of the Neural Cellular Automata model to two new tasks: a texture synthesis task and a feature visualisation task. Additionally, we show how training NCA approximates identification of a PDE within a family of PDEs often used to model reaction diffusion systems, evaluated in discretised time and space. We demonstrate remarkable qualities of the model: robustness, quick convergence, learning a qualitatively algorithmic solution, and relative invariance to the underlying computational implementation and manifold. These qualities suggest that models designed within the paradigm of self-organisation may be a promising approach to achieving more generalisable and robust artificial systems.

{
\small
\bibliographystyle{ieee_fullname}
\bibliography{egbib}
}

\end{document}